\documentclass[letterpaper]{article} 
\usepackage[submission]{aaai2026}  
\usepackage{times}  
\usepackage{helvet}  
\usepackage{courier}  
\usepackage[hyphens]{url}  
\usepackage{graphicx} 
\usepackage{natbib}  
\usepackage{caption} 
\usepackage{multirow}
\usepackage{algorithm}
\usepackage{algorithmic}
\usepackage{amsmath}
\usepackage{amsfonts}
\usepackage{float}
\usepackage{graphicx}
\usepackage{caption}
\usepackage{lipsum}
\usepackage{booktabs}

\usepackage{newfloat}
\usepackage{listings}
\usepackage{graphicx}
\usepackage{float}
\usepackage{subfig}
\usepackage{colortbl}
\usepackage{makecell}
\usepackage{multirow}
\usepackage{multicol}
\usepackage{placeins} 
\DeclareCaptionStyle{ruled}{labelfont=normalfont,labelsep=colon,strut=off} 
\floatstyle{ruled}
\newfloat{listing}{tb}{lst}{}
\floatname{listing}{Listing}
\title{Visual-CoG: Stage-Aware Reinforcement Learning with \\ Chain of Guidance for Text-to-Image Generation}
\author{
    Yaqi Li\equalcontrib,
    Peng Chen\equalcontrib,
    Mingyang Han\equalcontrib,
    Pi Bu\equalcontrib,\\
    Haoxiang Shi,
    Runzhou Zhao,
    Yang Yao, 
    Xuan Zhang, 
    Jun Song $^\dagger$,
    Bo Zheng
}

\affiliations{
    Alibaba Group

}

\usepackage{bibentry}
\usepackage{cuted}

\makeatletter
\def\showauthors@on{T}
\makeatother

\def\affiliations_{
  \textsuperscript{} Alibaba Group \\
  \{jiyan.lyq, zhaojun.cp, jingye.hmy, bupi.wj, jsong.sj\}@taobao.com
}
\begin{document}

\maketitle
\begin{strip}
  \centering
  \vspace*{-1.8cm}
  \includegraphics[width=\textwidth]{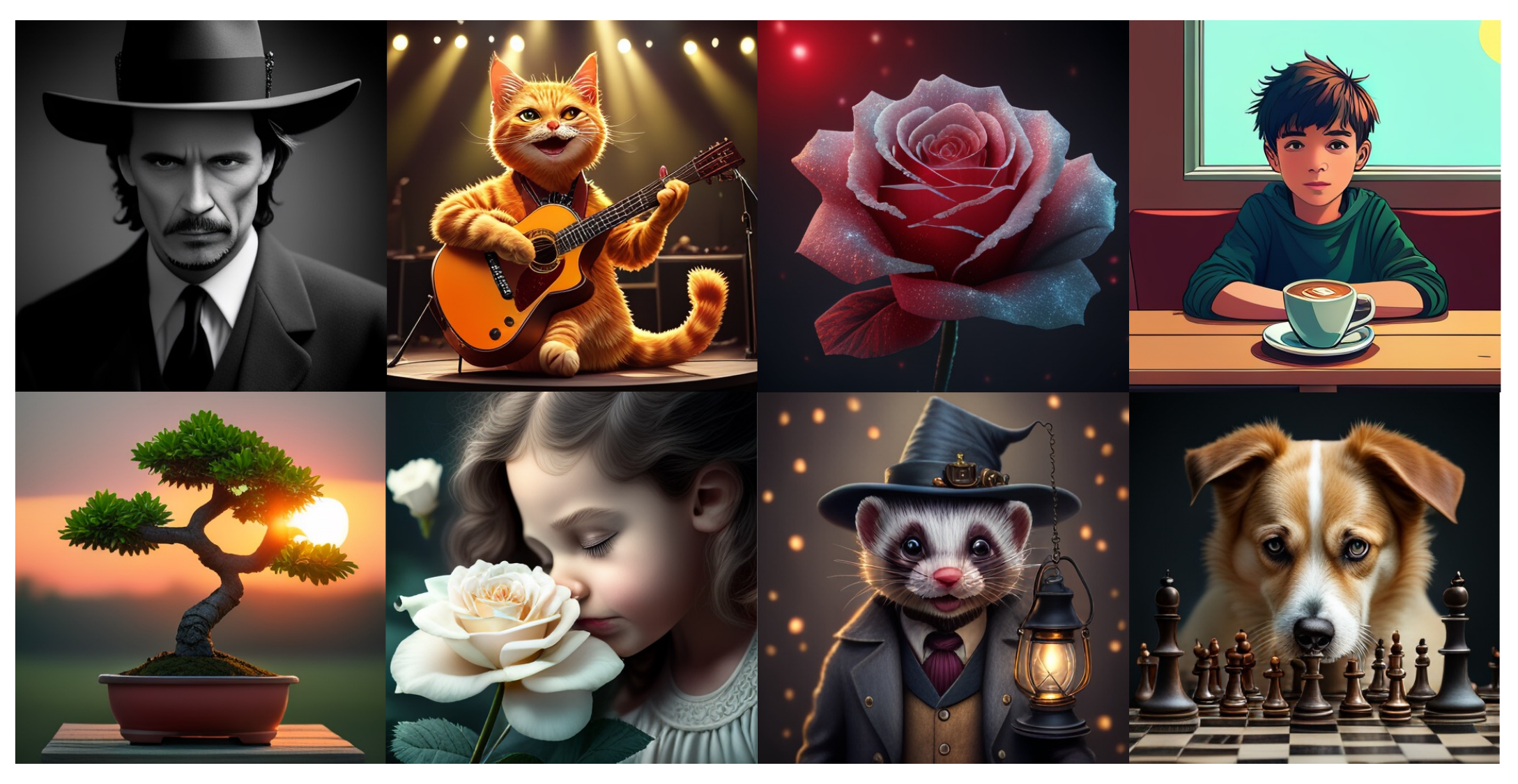}
  \vspace*{-0.8cm}
  \captionof{figure}{Visualization of the text-to-image generation results generated by our Visual-CoG.}
  \label{fig:vis1}
\end{strip}

\begin{abstract}
Despite the promising progress of recent autoregressive models in text-to-image (T2I) generation, their ability to handle multi-attribute and ambiguous prompts remains limited. 
To address these limitations, existing works have applied chain-of-thought (CoT) to enable stage-aware visual synthesis and employed reinforcement learning (RL) to improve reasoning capabilities.
However, most models provide reward signals only at the end of the generation stage. This monolithic final-only guidance makes it difficult to identify which stages contribute positively to the final outcome and may lead to suboptimal policies.
To tackle this issue, we propose a Visual-Chain of Guidance \textbf{(Visual-CoG)} paradigm consisting of three stages: semantic reasoning, process refining, and outcome evaluation, with stage-aware rewards providing immediate guidance throughout the image generation pipeline. 
We further construct a visual cognition benchmark, \textbf{VisCog-Bench}, which comprises four subtasks to evaluate the effectiveness of semantic reasoning.
Comprehensive evaluations on GenEval, T2I-CompBench, and the proposed VisCog-Bench show improvements of 15\%, 5\%, and 19\%, respectively, demonstrating the superior performance of the proposed Visual-CoG. We will release all the resources soon.


\end{abstract}
\section{Introduction}



Recent advancements in deep generative models have revolutionized visual content synthesis, with autoregressive models emerging as one of the leading paradigms \cite{li2024autoregressive, tian2024visual}. A recent trend in autoregressive models is to seamlessly integrate visual understanding and image generation into a single unified large model (ULM) built on multimodal large language models (MLLMs) \cite{xie2024show, tong2024metamorph}. This integration establishes a strong interdependence between vision understanding and generation, enabling ULMs to leverage the prior knowledge of MLLMs in interpreting complex instructions of image synthesis.
For instance, recent works \cite{liao2025imagegen, jiang2025t2i} utilize the reasoning capabilities of MLLMs to perform prior instruction planning and enriching, achieving a certain performance improvement.

\begin{figure*}[ht]
\centering
\includegraphics[width=0.95\textwidth]{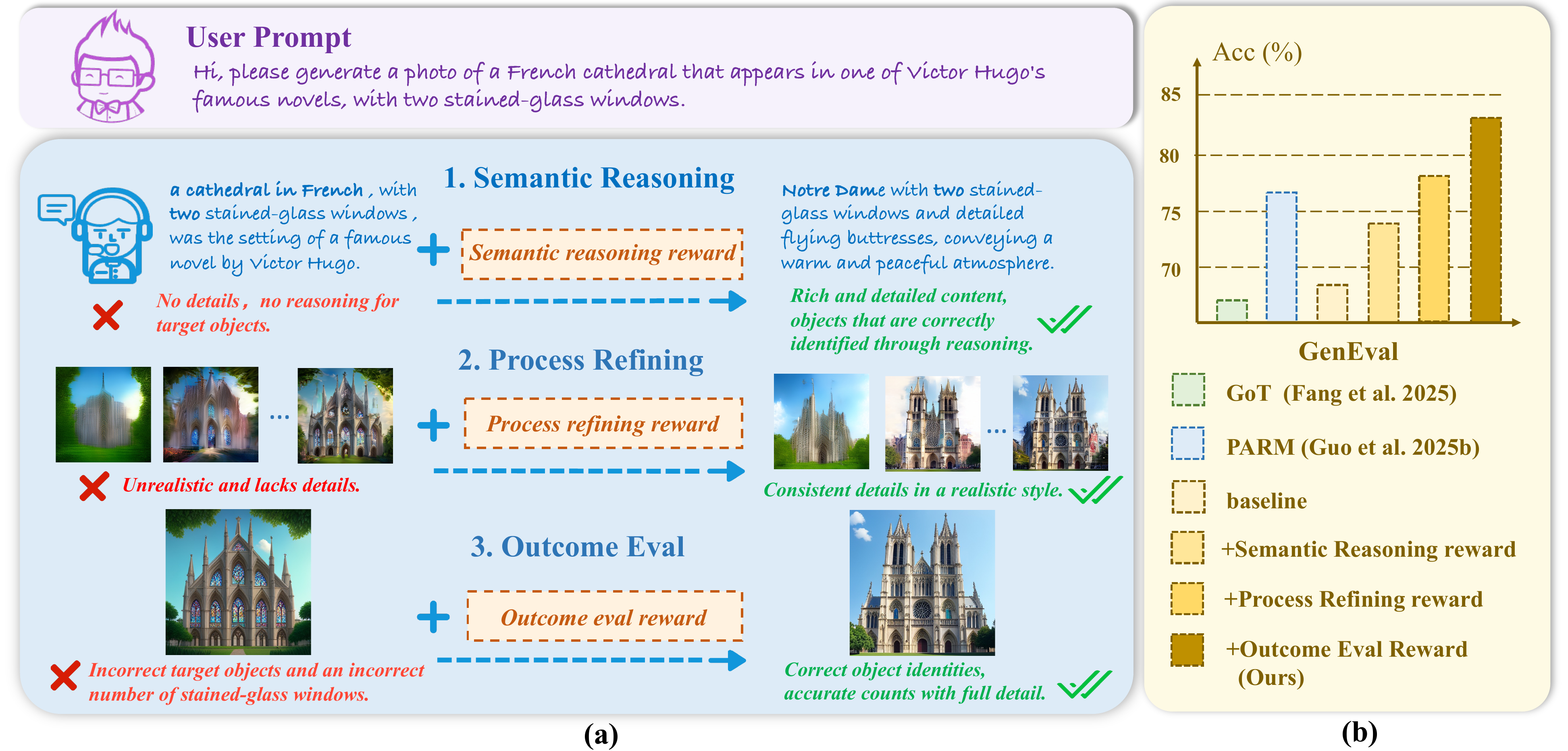} 
\vspace{-0.3cm}
\caption{Visualization of the image generation pipeline of Visual-Chain of Guidance \textbf{(Visual-CoG)}. (a) shows that with stage-aware guidance, the model is able to generate images that are more semantically aligned with prompts. (b) illustrates the effectiveness of the reward mechanisms across three stages of generation, as quantitatively evaluated on GenEval.}
\label{fig:vis2}
\vspace{-0.3cm}
\end{figure*}

While ULMs have seen significant advancements in recent years, they still face challenges in generating images that align with multi-attribute (\textit{e.g.}, color, counting, position) and ambiguous (\textit{e.g.}, the longest river in the world) prompts requiring reasoning \cite{ge2024seed, sun2024generative}, as demonstrated by failure cases in Fig.~\ref{fig:vis2}(a).
For multi-attribute prompts, we argue that applying chain-of-thought (CoT) to enable stage-aware visual synthesis is essential. In the same way, human artists typically follow an iterative creative process: they first understand the semantic concept, then progressively refine the object composition, and finally evaluate and analyze the result. This pipeline can be divided into three stages: semantic interpretation, progressive refinement, and result evaluation. For ambiguous prompts, we claim that only a model with strong reasoning capabilities can infer the true underlying intent. Inspired by the success of reinforcement learning (RL) in improving the reasoning performance of large language models (LLMs) \cite{shao2024deepseekmath, guo2025deepseek}, a few recent works \cite{yang2025hermesflow, duan2025got} have explored applying the RL paradigm to ULMs for instruction understanding or visual generation tasks. 
However, most existing works \cite{wang2025simplear} provide guidance (\textit{i.e.}, reward signals) only at the end of the generation process, ignoring the earlier stages such as semantic interpretation and progressive refinement. As shown in Fig.~\ref{fig:vis2} and Tab.~\ref{tab:tabRL}, this limitation makes it hard to determine which stage of the generation process contributes positively to the final outcome, potentially leading to ineffective or misleading policies.

In summary, we observe that (1) most ULMs struggle with multi-attribute and ambiguous prompts, and (2) recent works have explored CoT and RL to address these challenges yet typically provide guidance only at the final stage that ignores earlier ones. This monolithic final-only guidance tends to lead to suboptimal optimization. Therefore, a natural question arises: \textbf{Can we design stage-aware rewards for visual generation, enabling immediate feedback to guide the entire image generation pipeline?}

To achieve this, we propose a novel reinforcement learning-based framework named Visual-Chain of Guidance \textbf{(Visual-CoG)} for text-to-image generation. This framework formulates the visual generation as a chain of guidance by decoupling it into three distinct stages: \textbf{semantic reasoning}, \textbf{process refining}, and \textbf{outcome evaluation}, with immediate rewards provided at each stage as guidance. Specifically, the semantic reasoning reward refers to the discrepancy reward between the generated images based on the original prompt and the reasoning prompt obtained via semantic reasoning. The process refining reward evaluates the effect of the intermediate generation process using an auxiliary masked patch reconstruction task, while the outcome evaluation reward involves a rule-based, multidimensional assessment of the final output. 
Through the proposed paradigm, immediate reward signals can be obtained throughout the visual generation process, making it easier to identify the stages that require optimization. As shown in Fig.~\ref{fig:vis1}, the generated images demonstrate enhanced detail and diverse styles.

By conducting extensive experiments, Visual-CoG exhibits significant improvement on benchmarks including GenEval \cite{ghosh2023geneval} and T2I-CompBench \cite{huang2023t2i}. To further demonstrate the importance of semantic reasoning, we develop a comprehensive visual cognition benchmark called \textbf{VisCog-Bench}. It includes four subtasks, namely unusual position, unusual composition, unusual color, and reasoning tasks, to thoroughly evaluate a model's capability of interpreting implicit intentions.
Quantitative and qualitative analysis demonstrate that our method, guided by stage-aware rewards, enables the model to generate high-fidelity images for multi-attribute and ambiguous prompts, as illustrated in Fig.~\ref{fig:vis2}.

Our contributions are summarized as follows:
\begin{itemize}
    \item We propose a novel reinforcement learning-based framework for text-to-image generation, which implements a stage-aware Visual-Chain of Guidance \textbf{(Visual-CoG)} pipeline comprising semantic reasoning, process refining, and outcome evaluation stages, providing immediate feedback at each stage to guide optimization.
    \item We introduce a comprehensive visual cognition benchmark, \textbf{VisCog-Bench}, comprising four subtasks: unusual position, unusual composition, unusual color, and reasoning tasks, to thoroughly evaluate semantic reasoning ability in unusual and reasoning-demanding scenarios.
    \item Extensive experiments on GenEval, T2I-CompBench, and the proposed VisCog-Bench show that our method outperforms existing methods especially on multi-attribute and ambiguous prompts that require reasoning.
\end{itemize}

\section{Related Work}

\subsection{CoT and RL for Text-to-Image Models}
Recent works have explored chain-of-thought (CoT) and reinforcement learning (RL) to enhance text-to-image generation. 
PARM~\cite{guo2025can} enables CoT image generation with dynamic test-time verification and preference alignment, leading to improved generation performance.
GoT~\cite{fang2025got} leverages semantic-spatial reasoning in multimodal large language models (MLLMs) to improve visual generation and editing.
Additionally, SimpleAR~\cite{wang2025simplear} innovatively adopts the Group Relative Policy Optimization (GRPO)~\cite{shao2024deepseekmath} paradigm and uses the CLIP score as a reward signal for training the model.
However, these works primarily focus on evaluating only the final generated images, ignoring early generation stages.
In contrast, our Visual-CoG introduces stage-aware rewards that provide immediate feedback throughout the image generation, thereby enabling more effective policy learning and improved semantic alignment. 
\begin{figure*}[ht]
\centering
\includegraphics[width=0.94\textwidth]{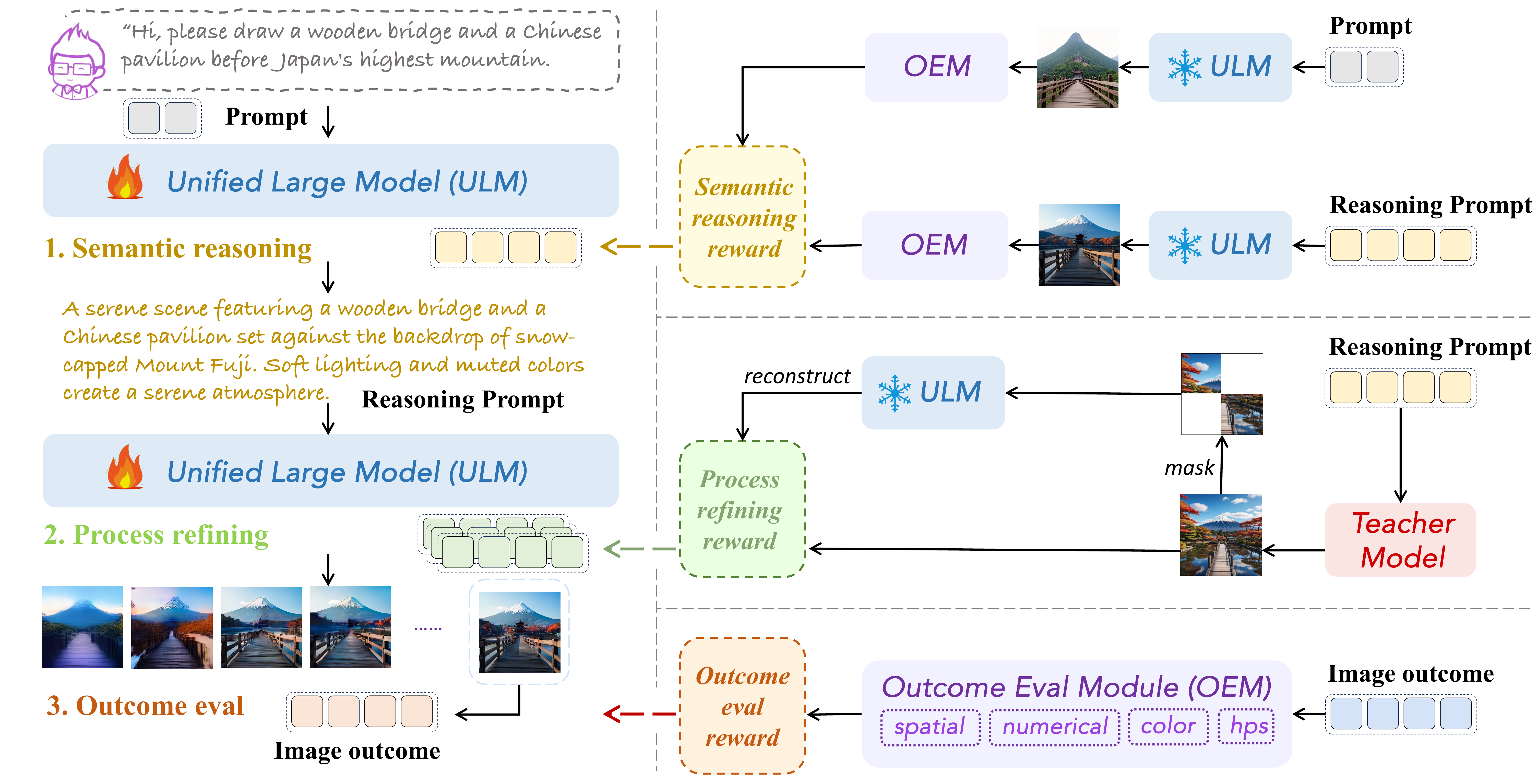} 
\vspace{-0.3cm}
\caption{An overview of Visual-CoG framework, which comprises three key stages: semantic reasoning, process refining, and outcome evaluation. Stage-aware rewards enable immediate feedback to guide the entire image generation pipeline.}
\vspace{-0.5cm}
\label{fig:vis3}
\end{figure*}

\subsection{Unified Large Models (ULMs)}

Recently, researchers have tried to apply the next-token-prediction paradigm to visual foundation models for unified visual understanding and generation.
One line of methods~\cite{sun2024generative, sun2023emu, ge2024seed} employs continuous embeddings to represent images, while relying on external image generation models (\textit{i.e.}, diffusion models) to complete the generation process. Additionally, Orthus~\cite{kou2024orthus} and Transfusion~\cite{zhou2024transfusion} explore continuous representations for unified modeling without the need for additional integration of an image generation model. 
In contrast, another line of methods unifies text and image modalities using discrete tokens, ensuring simplicity in the model structure. For example, EMU3~\cite{wang2024emu3} tokenizes images, text, and videos into a discrete space and builds a single transformer from scratch on a mixture of multimodal sequences. 
Moreover, some models such as Unified-IO2~\cite{lu2024unified} and Janus-Pro~\cite{chen2025janus} combine discrete and continuous representations via two different decoupled encoders for text and image generation respectively.

\section{Method}
\subsection{Preliminary}

\subsubsection{Mask Token Prediction} 
Mask-based generative modeling is a widely adopted technique in autoregressive image generation \cite{chang2022maskgit, xie2024show}, where the model learns to predict a set of masked tokens via bidirectional attention mechanisms. During inference, it initially generates all image tokens simultaneously and then iteratively refines the image conditioned on the previously generated tokens. The process can be formulated as follows:

\begin{equation}
\mathcal{L}_{\mathrm{MTP}}=\sum_{j} \log p_{\theta}\left(I_{j} \mid T_{1}, \cdots, T_{S}, I_{m}, I_{2}, \cdots, I_{m}, I_{N}, \right),
\label{denoise}
\end{equation}
where $p\left ( \cdot | \cdot \right )$ represents the conditional probability, parameterized by $\theta$. Let $\mathbf{T}=\left\{T_{1}, T_{2}, \cdots, T_{S}\right\}$ and $\mathbf{I}=\left\{I_{1}, I_{2}, I_{m}, \cdots, I_{N}\right\}$ denote text token sequence of length $S$ and image token sequence of length $N$ respectively. And $I_{m}$ refers to the masked image tokens that are to be predicted.
This strategy serves as the basis for the image refinement process in our chain-of-guidance framework.

\subsubsection{RL For Unified Large Models (ULMs)}
Reinforcement Learning (RL) seeks to learn an optimal policy $\pi_{\theta}$ that maximizes the reward $\mathbf{\mathcal{R}}$ return from interacting with an environment \cite{li2017deep}. For autoregressive ULMs, the state at time $t$ is the combination of the input $x$ (\textit{i.e.}, text or image) and the partially generated content $y<t$, and the action is the generation of the next token $y_{t}$.

To optimize this sequential generation process, policy gradient methods are adopted to directly maximize the expected return. The advantage function $A_{t}$ serves as a critical measure of how much an action outperforms the expected average performance. 
A widely adopted paradigm in RL is formalized as follows:
\begin{equation}
\mathcal{L}(\theta) = - \mathbb{E}_{t} \left [ \min\left ( \gamma_{t}\left ( \theta  \right ) A_{t}, \text{clip}\left ( \gamma_{t}\left ( \theta  \right ), 1-\epsilon , 1+\epsilon \right ) A_{t} \right ) \right ],
\label{loss}
\end{equation}

\begin{equation}
\gamma_{t}\left ( \theta  \right ) =\frac{\pi_{\theta}(y_t \mid \mathbf{y}_{<t})}{\pi_{\theta_{\text{old}}}(y_t \mid \mathbf{y}_{<t})},
\label{importance_sample}
\end{equation}
where $(x,y)$ denotes a given input-output pair, $\epsilon$ is a clipping parameter, $A_{t}$ is the normalized reward for a group of $G$ samples $\mathbf{\mathcal{R}} =\{\mathcal{R}_1, \mathcal{R}_2, \dots, \mathcal{R}_G\}$ computed as $A_{t}^{i} = \frac{\mathcal{R}_i - \text{mean}(\mathbf{\mathcal{R}})}{\text{std}(\mathbf{\mathcal{R}})}$ and $A_{t} = \frac{1}{G} \sum_{i=1}^{G} A_{t}^{i}$.

\subsection{Visual-Chain of Guidance}
Drawing inspiration from the iterative creative process of human artists, which involves first understanding the semantic concept, progressively refining the object composition, and finally evaluating and analyzing the result, we introduce a Visual-Chain of Guidance (\textbf{Visual-CoG}) framework with stage-aware rewards for image generation. The pipeline includes \textbf{(1) Semantic Reasoning, (2) Process Refining} and \textbf{(3) Outcome Evaluation}, as elaborated below.

\subsubsection{Semantic Reasoning}
Initial semantic reasoning is a stage prior to image generation, which involves reasoning about the underlying intention of the instruction and planning the scene layout. In unusual or reasoning-demanding scenarios, ULMs leverage their inherent ability to interpret visual inputs to concretize and clarify the generation instructions.

Specifically, this stage begins with semantic reasoning via a language modeling task performed by a language model $\mathcal{M}_{lm}$ to generate a reasoning prompt $\mathcal{P'}$ based on the original prompt $\mathcal{P}$. $\mathcal{P}$ and $\mathcal{P'}$ are then used as inputs of image generation task $\mathcal{M}_{t2i}$ under the same random seed, resulting in two generated images $\mathcal{I}$ and $\mathcal{I'}$. An outcome evaluation reward $\mathcal{R}_{o}$ then computes scores for both images $\mathcal{I}$ and $\mathcal{I'}$, which are designed to assess image-text alignment comprehensively (see \textit{Outcome Evaluation} for details). The difference between these scores is defined as the semantic reasoning reward $\mathcal{R}_{r}$, which reflects the effectiveness of the reasoning process. Intuitively, a higher score indicates better reasoning performance, thus serving as an intermediate feedback to optimize the model. The semantic reasoning reward is as follows:

\begin{equation}
\mathcal{R}_{r}=\mathcal{R}_{o}(\mathcal{M}_{t2i}(\mathcal{P'})) - \mathcal{R}_{o}(\mathcal{M}_{t2i}(\mathcal{P})).
\label{denoise3}
\end{equation}

\begin{table*}[!htbp]
\vspace{-0.4em}
\caption{Comparison to the state-of-the-art diffusion models and autoregressive models on GenEval.}
\vspace{-0.3cm}
\centering
\resizebox{0.9\textwidth}{!}{
\begin{tabular}{ l| c c c c c c |c}
\toprule
 Method & Single Obj. & Two Obj. & Counting & Colors & Position & Color Attri. & Overall\\
\midrule
\rowcolor{gray!30} \textit{Diffusion Models} & & & & & & &\\

Stable v1.5 \cite{rombach2022high}  & 97.13 & 38.21 & 35.88 & 76.41 & 4.55 & 6.97 & 43.19\\
SD-XL-base-1.0 \cite{podell2023sdxl}   & 98.34 & 74.14 & 39.21 & 85.36 & 15.09 & 23.11 & 55.87 \\

PixArt-$\alpha$ \cite{chen2023pixart} & 98.05 & 50.31 & 44.67 & 80.12 & 8.55 & 7.21 & 48.15 \\
SD-3 (d=24) \cite{esser2024scaling} & 98.05 & 74.55 & 63.05 & 67.12 & 34.09 & 36.29 & 62.19 \\
\midrule
\rowcolor{gray!30} \textit{Autoregressive Models} & & & & & & &\\
SEED-X \cite{ge2024seed} & 97.01 & 58.24 & 26.13 & 80.27 & 19.51 & 14.90 & 49.34 \\	
PARM \cite{guo2025can}  & 99.21 & 85.64 & 67.15 & 84.02 & 65.83 & 64.04 & 77.64 \\	
Janus-Pro-7B \cite{chen2025janus}  & 99.03 & 88.98 & 58.75 & \textbf{90.02} & 78.89 & \textbf{66.12} & 80.29 \\	
Show-o (Baseline) \cite{xie2024show}  & 98.11 & 80.04 & 66.25 & 84.17 & 31.03 & 50.14 & 68.29 \\
\hline
\textbf{Visual-CoG (Ours)} & \textbf{99.95} & \textbf{92.68} & \textbf{80.94} & 85.11& \textbf{79.00} & 65.50& \textbf{83.86} \\
\bottomrule
\end{tabular}
}
\vspace{-0.5cm}
\label{tab:tab1}
\end{table*}

\subsubsection{Process Refining}
Process Refining refers to the iterative and adaptive refinement of intermediate generation processes via mask token prediction.
In contrast to methods that assess only the final output, we propose an immediate reward mechanism for the image refinement process, which provides immediate feedback during the generation process. 
As discussed in the section \textit{preliminary}, at each intermediate generation step, the model generates an image conditioned on a masked input under a given masking probability. The resulting output is then used as input for the next step, thus forming an iterative masked patch reconstruction process. 
Intuitively, a simple way to reward the generation process would be to evaluate the intermediate images directly. However, those intermediate images often suffer from blurriness or are incomplete, thus explicit image-based evaluation using detection methods is not suitable for providing reliable feedback during the generation process.

Instead, given the characteristics of the generation process, we perform a masked patch reconstruction task to evaluate the generation process in a straightforward yet effective way. Since it would be meaningless for the policy model to recover the masked image generated by itself, we introduce a teacher model $\tau$ to provide a preferred distribution. The model is expected to reconstruct this distribution under the corresponding masked condition.

Specifically, we define a process refining reward at step $t_m$ as the discrepancy between the preferred distribution $p_{\tau} = \pi_{\tau}(y_{t_{m}} \mid \mathbf{y}{<t_{m}})$ and the one reconstructed by the policy model $p_{\theta} = \pi_{\theta}(y_{t_{m}} \mid \mathbf{y}{<t_{m}})$. The reward $\mathcal{R}_{p}$ serves as an indicator of how well the model performs during the current generation process under the given masked condition. $\mathcal{R}_{p}$ is computed as follows:
\begin{equation}
\mathcal{R}_p = \exp \left( \left\lVert  \mathcal{G}(p_\theta)  -  \mathcal{G}(p_\tau)  \right\rVert^p \right),
\label{denoise3}
\end{equation}
where $\mathcal{G}$ converts the distribution to an image. The process refining reward enables more effective intermediate strategy adjustments, leading to higher-quality outputs.

\subsubsection{Outcome Evaluation}

Inspired by the rule-based reward mechanism in DeepSeek-R1~\cite{guo2025deepseek}, we design a rule-based reward framework to evaluate the quality of generated images. Specifically, we use an open-vocabulary object detector $\mathcal{D}$ with textual queries to determine whether the target objects are present in the generated image. Based on this detection, we then evaluate the consistency of spatial, counting, and color attributes using predefined rules. 

To assess \textbf{spatial consistency}, we employ a spatial validator $ \mathcal{E}_{s}$ to infer the relative positions (\textit{e.g.}, left of) of objects based on the 2D coordinates of their bounding box centroids. The inferred spatial relationships are then compared against the explicit spatial constraints specified in the prompt. Let $\mathcal{S}$ denote the set of object pairs $(i,j)$ with defined spatial relations, where $\left | \mathcal{S} \right | =N_s$. For each pair $(i,j)\in \mathcal{S} $, let ${y}_{(i,j)}^*$ represent the corresponding ground-truth relationship. The spatial reward $\mathcal{R}_{s}$ is computed as follows:

\begin{equation}
\mathcal{R}_{s} =\frac{1}{N_s} \sum_{(i,j) \in \mathcal{S}} \mathbb{I} \left( \mathcal{E}_{s}\left ( \mathcal{D}\left ( \mathcal{I}_{\text{e}},i \right ), \mathcal{D}\left ( \mathcal{I}_{\text{e}},j \right ) \right )  = {y}_{(i,j)}^* \right),
\label{denoise3}
\end{equation}
For \textbf{counting consistency}, we check whether the number of generated objects matches the target quantity $y_{i_n}^{*}$ from the prompt. Here, $N_n$ refers to the number of objects with numerical attributes in image $\mathcal{I}_{\text{e}}$, and $\mathcal{E}_{n}$ is used to convert the detection results of object $i$ into a specific number. To provide a smooth reward, we design an exponential function with a scaling factor that penalizes larger deviations more severely as follows:

\begin{equation}
\mathcal{R}_{n} = \frac{1}{N_n} \sum_{i=1}^{N_n} \text{exp}\left ( \frac{ \left | \mathcal{E}_{n}\left ( \mathcal{D}\left ( \mathcal{I}_{\text{e}},i \right ) \right )  - y_{i_n}^* \right | }{\tau }  \right ),
\label{denoise3}
\end{equation}

To evaluate \textbf{color consistency}, we utilize a zero-shot object-centric classification approach based on a pre-trained vision-language model (\textit{i.e.}, CLIP). For each detected object $i$, a post-processing step $\mathcal{E}_{c}$ is applied, which involves extracting its bounding box and refining it with a segmentation mask to isolate the foreground. A set of textual prompts in the format “a photo of a [color] [object]” is then generated for each candidate color $y$ in the predefined color list $\mathcal{Y}$. The vision-language model computes similarity scores between the processed object $i$ and each prompt $\mathcal{P}_{y,i}$, and the color with the highest score is selected as the predicted label. This prediction is compared against the ground-truth label $y_{i_c}^*$ to assess consistency. Here, $N_c$ denotes the number of objects with specified color attributes.

\begin{equation}
\mathcal{R}_{c} = \frac{1}{N_c} \sum_{i=1}^{N_c} \mathbb{I} \left( \arg\max_{y \in \mathcal{Y}} \text{sim}(\mathcal{E}_{c}\left ( \mathcal{D}\left ( \mathcal{I}_{\text{e}},i \right ) \right ), \mathcal{P}_{y,i}) = y_{i_{c}}^{*} \right),
\label{denoise3}
\end{equation}

In addition to the attribute-wise rewards at the local level, it is essential to evaluate the \textbf{overall aesthetic quality and alignment} from a global perspective. Therefore, we employ the HPS model \cite{wu2023human} as an additional reward model to provide a holistic score $\mathcal{R}_{h}$ for the image. 

The reward for the image outcome evaluation $\mathcal{R}_{o}$ is as follows. Note that $\mathcal{R}_{o}$ means $\mathcal{R}_{o}(\mathcal{I}_{e})$ by default for simplicity.
\begin{equation}
\mathcal{R}_{o}= \mathcal{R}_{n} + \mathcal{R}_{c} + \mathcal{R}_{s} + \mathcal{R}_{h},
\label{denoise3}
\end{equation}

The overall reward $\mathcal{R}$ is the sum of the three stages and the loss is defined in Eq. \ref{loss}:
\begin{equation}
\mathcal{R}= \mathcal{R}_{r} + \mathcal{R}_{p} + \mathcal{R}_{o},
\label{denoise3}
\end{equation}

\begin{table*}[tbp]
\vspace{-0.4em}
\caption{Comparison to the state-of-the-art diffusion models and autoregressive models on T2I-CompBench.}
\vspace{-0.3cm}
\centering

\resizebox{0.8\textwidth}{!}{
\begin{tabular}{ l| c c c c c c}
\toprule
 Method & Color & Shape & Texture & Spatial & Non-Spatial & Complex \\
\midrule
\rowcolor{gray!30} \textit{Diffusion Models} & & & & & & \\

Stable v1.5 \cite{rombach2022high}  & 37.58 & 37.13 & 41.86 & 11.65 & 31.12 & 30.47 \\
SD-XL-base-1.0 \cite{podell2023sdxl}  & 58.79 & 46.87 & 52.99 & 21.31 & 31.19 & 32.37 \\
Composable v2 \cite{liu2022compositional} & 40.63 & 32.99 & 36.45 & 8.00 & 31.12 & 29.80 \\
PixArt-$\alpha$ \cite{chen2023pixart} & 66.90 & 49.27 & 64.77 & 20.64 & 31.97 & 34.33 \\					
\midrule
\rowcolor{gray!30} \textit{Autoregressive Models} & & & & & & \\
PARM \cite{guo2025can}& 75.21 & 56.32 & 66.04 & 29.12 & \textbf{31.44} & 36.81 \\
Janus-Pro-7B \cite{chen2025janus} & 63.59 & 35.28 & 49.36 & 20.61 & 30.85 & 34.31 \\
Show-o (Baseline) \cite{xie2024show}  & 68.45 & 49.61 & 65.34 & 38.12 & 30.06 & 34.67 \\
\hline
\textbf{Visual-CoG (Ours)} & \textbf{78.92} & \textbf{57.49} & \textbf{67.85} & \textbf{43.71} & 30.90 & \textbf{36.84} \\
\bottomrule
\end{tabular}
}
\vspace{-0.5cm}
\label{tab:tab2}
\end{table*}

\vspace{-0.5cm}
\section{Experiments}
\subsection{Experimental Setup}
\subsubsection{Implementation Details}
Our training dataset consists of text prompts generated using Qwen3 \cite{yang2025qwen3} based on several templates covering the 80 COCO \cite{lin2014microsoft} object classes (see \textit{Appendix} for details). These prompts are further processed by Qwen3 to extract objects and their associated attributes (\textit{e.g.}, color, position and counting) for reward computation. A total of 9,368 prompts are generated in this manner for training. We use Show-o \cite{xie2024show} as our base model and train it with a learning rate of 1e-6. For the reward modeling component, we adopt HPS \cite{wu2023human} as the human preference model and GroundingDINO \cite{liu2024grounding} as the object detector. The proposed model is implemented using PyTorch and trained on NVIDIA H20 GPUs.

\subsubsection{Evaluation Metrics}
We evaluate the effectiveness of our method on the GenEval \cite{ghosh2023geneval} and T2I-CompBench \cite{huang2023t2i} benchmarks, which are widely adopted for evaluating text-image alignment from the attribute-wise perspective. We adopt the recommended protocols for both benchmarks. 

\subsubsection{VisCog-Bench}
To further verify the effectiveness of reasoning capabilities, we propose a novel visual cognition benchmark, \textbf{VisCog-Bench}, comprising four subtasks: unusual position, unusual composition, unusual color, and reasoning tasks. The benchmark consists of 20, 20, 20, and 40 prompts for each task respectively, totaling 100 prompts. Note that 4 images are generated for each prompt for evaluation to avoid randomness. To ensure a more comprehensive evaluation, we assess the generated images using both automated metrics and human evaluations. Specifically, the prompts for unusual position, unusual composition, and unusual color are selected from GenEval, and we adopt the corresponding metrics of GenEval. For the reasoning tasks, we use Qwen3 to generate prompts that require reasoning based on common knowledge, and the resulting images are automatically evaluated using Qwen2.5-VL \cite{bai2025qwen2}. Additionally, we conduct a user study as part of the evaluator-based evaluation (see \textit{Appendix} for details).


\subsection{Quantitative results}

Tab.~\ref{tab:tab1} and Tab.~\ref{tab:tab2} present a comprehensive comparison between the proposed method and other state-of-the-art models across the visual generation benchmarks GenEval and T2I-CompBench, covering both diffusion-based methods such as SD-3 \cite{esser2024scaling}, PixArt-$\alpha$ \cite{chen2023pixart} and autoregressive-based models such as SEED-X \cite{ge2024seed}, Janus-Pro \cite{chen2025janus}. 
Notably, our method achieves substantial improvements over the baseline model Show-o \cite{xie2024show}, with an average enhancement of 15.57\% across all metrics on GenEval. It consistently outperforms both diffusion-based and autoregressive-based approaches, achieving a 13.79\% improvement in the counting subtask compared to previous state-of-the-art results. 

Similarly, on T2I-CompBench, our method achieves strong performance across various tasks, obtaining the highest scores of 78.92\% in \textit{Color} and 43.71\% in \textit{Spatial}. These improvements are partly attributed to the semantic reasoning stage, which enables global-level planning and reasoning. And subsequent image process refining and output evaluation ensure faithful execution of the initial design intent, thereby guaranteeing accurate and coherent final results.
Moreover, autoregressive models consistently outperform diffusion models in prompt interpretation, leading to more accurate text-image alignment across all attributes in both GenEval and T2I-CompBench benchmarks.

Additionally, Tab.~\ref{tab:tabRL} displays a comparison with autoregressive models that employ reinforcement learning (RL) or chain-of-thought (CoT)-based techniques. It is evident that the proposed Visual-CoG significantly outperforms others. This suggests that while these methods incorporate RL and CoT techniques to enable a stage-by-stage generation, they typically provide guidance only at the final stage, ignoring earlier stages and resulting in suboptimal performance. 

\begin{table}[tbp]

\caption{Comparison to outstanding autoregressive models with RL or CoT on GenEval.}
\vspace{-0.3cm}
\centering
\resizebox{1.0\columnwidth}{!}{
\begin{tabular}{c | c c c |c}
\toprule
Method & Counting  & Position & Color Attri. & Overall  \\
\midrule
VARGPT-v1.1 \cite{zhuang2025vargpt} & 48.00 & 13.00 & 21.00 & 53.00 \\
SimpleAR \cite{wang2025simplear} & - & 28.00 & 45.00 & 63.00 \\
GoT \cite{fang2025got} & 67.00 & 34.00 & 27.00 & 64.00\\
RePrompt \cite{wu2025reprompt}& 77.00 & 62.00 & 49.00 & 76.00 \\
PARM \cite{guo2025can}& 67.15 & 65.83 & 64.04 & 77.64 \\
\textbf{Visual-CoG (Ours)}&80.94& 79.00 & 65.50 & 83.86\\
\bottomrule
\end{tabular}
}
\label{tab:tabRL}
\end{table}
\vspace{-0.3cm}

\begin{table}[tbp]
\vspace{-0.4em}
\caption{Effect of rewards on different stages.}
\vspace{-0.3cm}
\centering
\resizebox{.95\columnwidth}{!}{
\begin{tabular}{c c c| c c c |c}
\toprule
$\mathcal{R}_{r}$ & $\mathcal{R}_{p}$ & $\mathcal{R}_{o}$ & Counting  & Position & Color Attri. & Overall  \\
\midrule
 - & - & - & 66.25 & 31.03 & 50.14 & 68.29\\
 \checkmark & \checkmark & - & 76.87 & 76.51 & 55.97 & 78.12 \\
 \checkmark & - & \checkmark & 72.98 & 75.09 & 56.59 & 76.23\\
 - & \checkmark  & \checkmark & 78.13 & 72.01 & 62.88 & 80.11 \\
\checkmark  & \checkmark & \checkmark &80.94& 79.00 & 65.50 & 83.86\\
\bottomrule
\end{tabular}
}
\vspace{-0.3cm}
\label{tab:tab3}
\end{table}


\begin{table}[tbp]
\vspace{-0.4em}
\caption{VisCog-Bench results including four subtasks.}
\vspace{-0.3cm}
\centering
\resizebox{.98\columnwidth}{!}{
\begin{tabular}{l| c c c c |c c}
\toprule
\multirow{2}{*}{Method}  & \multicolumn{3}{c}{Unusual}  & \multirow{2}{*}{Reasoning}  & \multirow{2}{*}{Overall} & \multirow{2}{*}{Human Eval} \\
\cline{2-4}
  & Color& Pos. & Comp. &  & \\
\midrule
Stable v1.5 & 11.25 & 3.75 & 31.25 & 38.98 & 21.30 & 23.78\\
PixArt-$\alpha$ & 18.75 & 7.50 & 42.50 & 40.78 & 27.38 & 25.98\\
Show-o & 62.25 & 28.75 & 76.75 & 64.54 & 58.07 & 55.21 \\
\hline
Ours w/o $\mathcal{R}_{r}$ & 64.50 & 49.75 & 79.25 & 67.89 & 65.34 & 61.30\\
Ours & 70.00 & 77.50 & 90.00 & 72.50 & 77.50 & 78.55 \\
\bottomrule
\end{tabular}
}
\vspace{-0.3cm}
\label{tab:tab4}
\end{table}

\begin{figure*}[ht]
\centering
\includegraphics[width=0.97\textwidth]{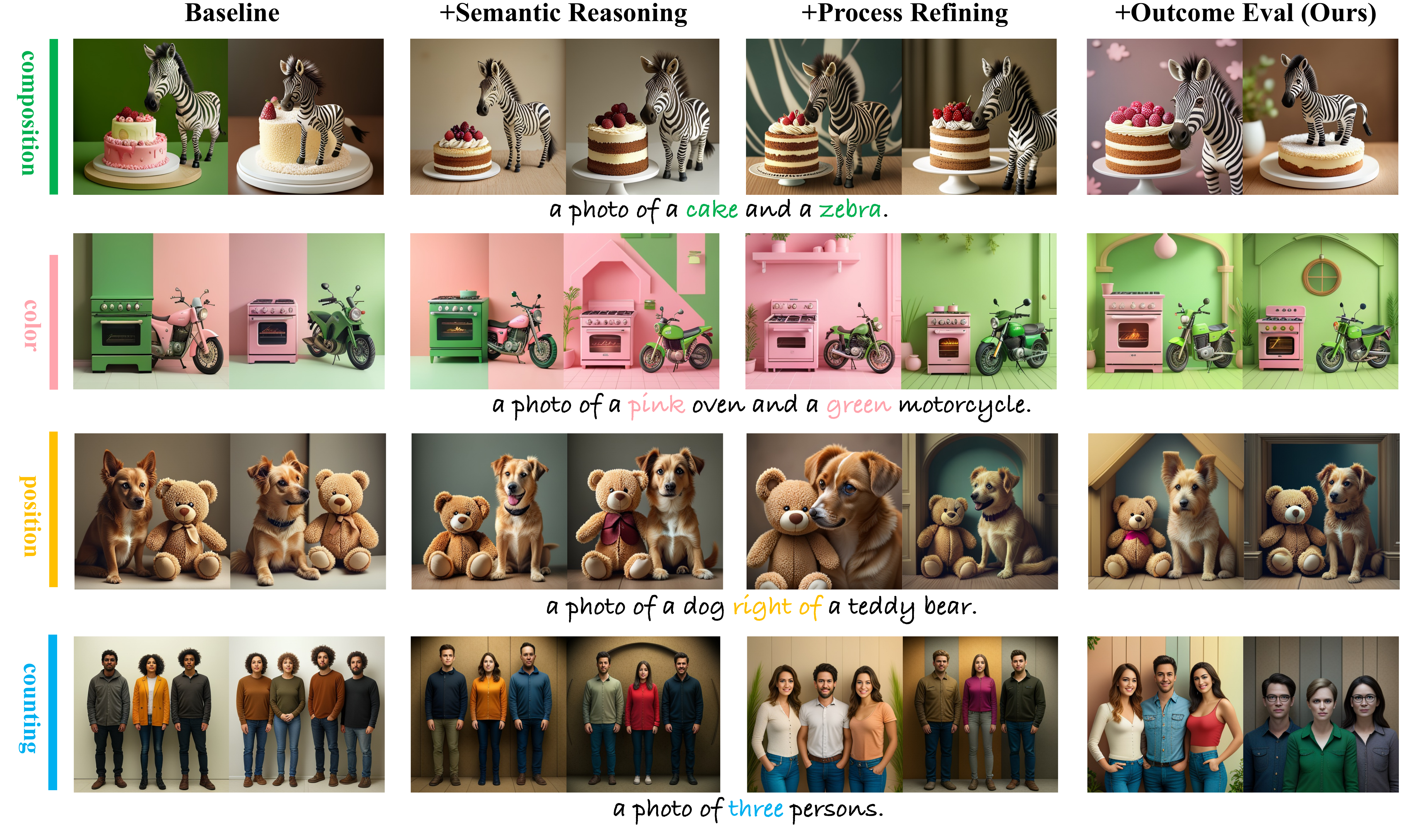} 
\vspace{-0.5cm}
\caption{Qualitative results from GenEval for various attributes such as composition, color, position and counting. }
\label{fig:vis4}
\vspace{-0.3cm}
\end{figure*}

\subsection{Ablation studies}
To explore the contribution of the semantic reasoning reward $\mathcal{R}_{r}$, process refining reward $\mathcal{R}_{p}$, and outcome evaluation reward $\mathcal{R}_{o}$, we conduct ablation studies on the GenEval benchmark, across counting, position, and color metrics. As shown in Tab.~\ref{tab:tab3}, all rewards consistently enhance alignment performance. Specifically, $\mathcal{R}_{r}$ leads to a considerable improvement of 6.99\% in position metrics, while $\mathcal{R}_{p}$ and $\mathcal{R}_{o}$ significantly enhance counting and color metrics, with improvements of 7.96\% and 9.53\%, respectively. This can be attributed to $\mathcal{R}_{r}$ providing prior planning for scene construction and object layout, enabling accurate generation of position-related scenarios. Meanwhile, $\mathcal{R}_{p}$ ensures consistency in local details (\textit{e.g.}, color and counting relationships) during refinement, as the basis for subsequent outcome evaluation. $\mathcal{R}_{o}$ offers direct feedback on outcomes, explicitly evaluating attributes such as color. The combination of these rewards further enhances overall semantic alignment.

As shown in Fig.~\ref{fig:vis4}, the generated images under different settings: using only $\mathcal{R}_{r}$ (2nd column), with $\mathcal{R}_{r}$ and $\mathcal{R}_{p}$ (3rd column), and with $\mathcal{R}_{r}$, $\mathcal{R}_{p}$, and $\mathcal{R}_{o}$ (4th column). These settings are assessed across on composition, color, position, and counting metrics.
In the first and third rows, the images generated by the baseline misrepresent the composition or positional relationships between dogs and bears respectively. Semantic reasoning corrects these issues through accurate object layout planning. As shown in the second row, process refining enhances the details of the generated images at the local level. This ensures that the pink oven and the green motorcycle merge seamlessly, with no visual discontinuity. Additionally, images generated with outcome evaluation reward demonstrate high aesthetic quality and diversity. For instance, the image in the last column features a more detailed background and natural atmosphere.

\subsection{Semantic Reasoning Analysis}
We conduct experiments on the proposed VisCog-Bench to verify the effectiveness of reasoning capabilities in the context of unusual or reasoning-demanding prompts. Generating accurate images for VisCog-Bench requires the model to infer the described objects and scenarios, emphasizing the role of early semantic reasoning.

\begin{figure}[t]
\centering
\includegraphics[width=1\columnwidth]{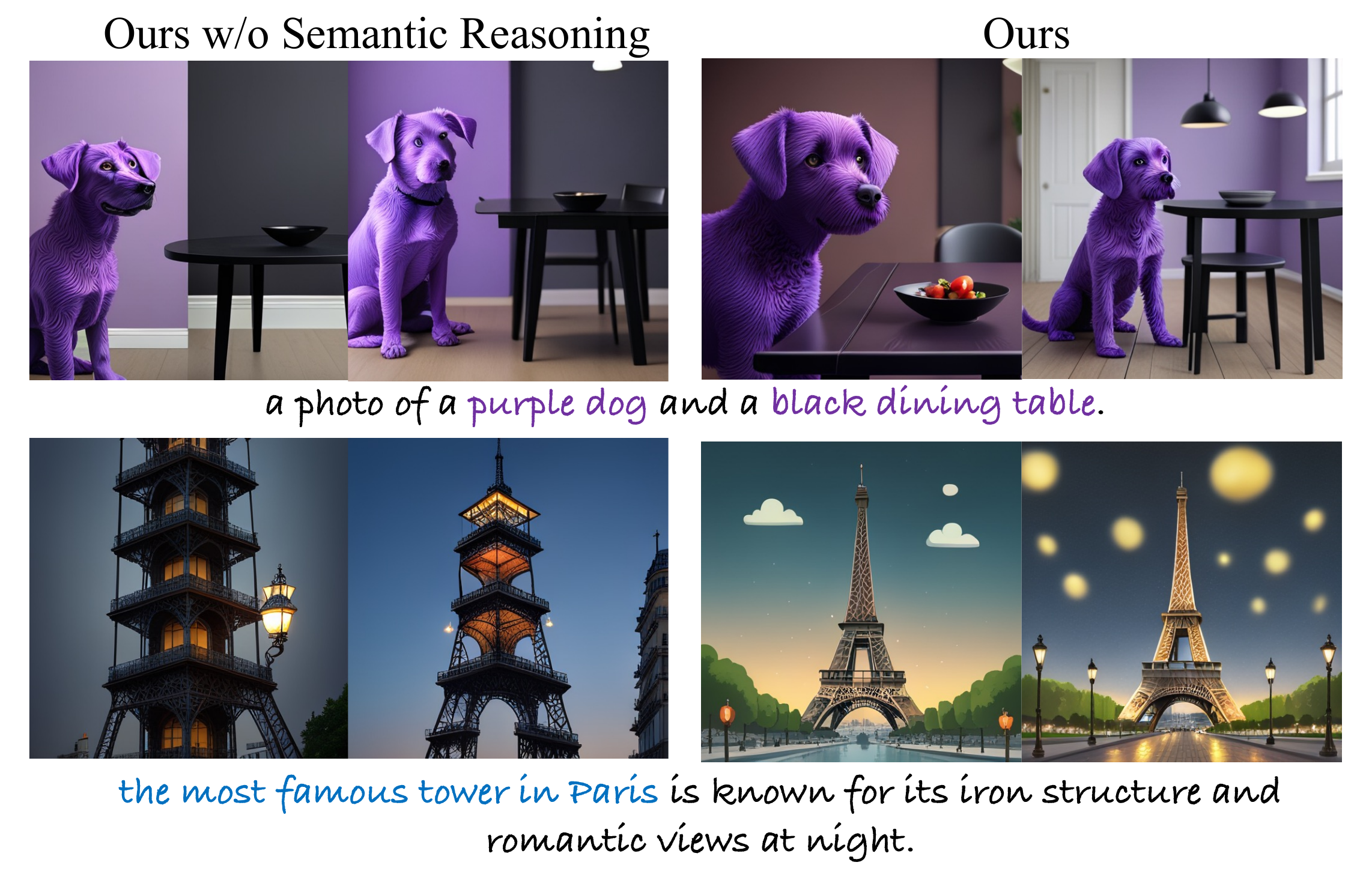} 
\vspace{-0.3cm}
\caption{Qualitative results of image generation for prompts involving unusual colors (1st row) and common-sense reasoning (2nd row), with and without semantic reasoning. }
\vspace{-0.5cm}
\label{fig:vis5}
\end{figure}
As demonstrated in Tab.~\ref{tab:tab4}, semantic reasoning leads to significant improvements for unusual or reasoning-demanding prompts, with gains of 12.16\% and 17.35\% in automated and human evaluation respectively. This is due to semantic reasoning enabling better interpretation of unusual or ambiguous prompts. As shown in Fig.~\ref{fig:vis5}, for unusual composition and color (1st row), models without semantic reasoning tend to generate disconnected scenarios (\textit{e.g.}, placing the dog and dinner table in separate regions of the image). In contrast, semantic reasoning enables prompt interpretation, leading to more detailed descriptions such as ``a dog with purple fur beside a dinner table in the house," thus producing more coherent and specific visual outputs. For reasoning tasks (2nd row), semantic reasoning helps interpret ambiguous concepts ``The most famous tower in Paris is the Eiffel Tower," enabling accurate image generation.

\section{Conclusion}
In this work, we propose a novel Visual-Chain of Guidance (\textbf{Visual-CoG}), which consists of three stages: semantic reasoning, process refining, and outcome evaluation, with stage-aware rewards providing immediate guidance throughout the image generation pipeline, particularly for multi-attribute and ambiguous prompts. Additionally, we introduce a visual cognition benchmark, \textbf{VisCog-Bench}, comprising four subtasks to evaluate the effectiveness of semantic reasoning. Extensive experiments show that the proposed method achieves outstanding performance across multiple benchmarks with enhanced reasoning capabilities.

\bigskip
\noindent 
\newpage
\clearpage
\bibliography{aaai2026}

\newpage
\clearpage
\newcommand{\answerYes}[1][]{\textcolor{blue}{[Yes] #1}}
\newcommand{\answerNo}[1][]{\textcolor{orange}{[No] #1}}
\newcommand{\answerNA}[1][]{\textcolor{gray}{[NA] #1}}
\renewcommand\thesubsection{\arabic{subsection}}
\renewcommand\labelenumi{\thesubsection.\arabic{enumi}}

\newcounter{checksubsection}
\newcounter{checkitem}[checksubsection]

\newcommand{\checksubsection}[1]{%
  \refstepcounter{checksubsection}%
  \paragraph{\arabic{checksubsection}. #1}%
  \setcounter{checkitem}{0}%
}

\newcommand{\checkitem}{%
  \refstepcounter{checkitem}%
  \item[\arabic{checksubsection}.\arabic{checkitem}.]%
}
\newcommand{\question}[2]{\normalcolor\checkitem #1 #2 \color{blue}}
\newcommand{\ifyespoints}[1]{\makebox[0pt][l]{\hspace{-15pt}\normalcolor #1}}
\section*{Reproduction Checklist}
\vspace{1em}
\hrule
\vspace{1em}

\checksubsection{General Paper Structure}
\begin{itemize}

\question{Includes a conceptual outline and/or pseudocode description of AI methods introduced}{(yes/partial/no/NA)}
yes

\question{Clearly delineates statements that are opinions, hypothesis, and speculation from objective facts and results}{(yes/no)}
yes

\question{Provides well-marked pedagogical references for less-familiar readers to gain background necessary to replicate the paper}{(yes/no)}
yes

\end{itemize}
\checksubsection{Theoretical Contributions}
\begin{itemize}

\question{Does this paper make theoretical contributions?}{(yes/no)}
yes

	\ifyespoints{\vspace{1.2em}If yes, please address the following points:}
        \begin{itemize}
	
	\question{All assumptions and restrictions are stated clearly and formally}{(yes/partial/no)}
	yes

	\question{All novel claims are stated formally (e.g., in theorem statements)}{(yes/partial/no)}
	yes

	\question{Proofs of all novel claims are included}{(yes/partial/no)}
	yes

	\question{Proof sketches or intuitions are given for complex and/or novel results}{(yes/partial/no)}
	yes

	\question{Appropriate citations to theoretical tools used are given}{(yes/partial/no)}
	yes

	\question{All theoretical claims are demonstrated empirically to hold}{(yes/partial/no/NA)}
	yes

	\question{All experimental code used to eliminate or disprove claims is included}{(yes/no/NA)}
	yes
	
	\end{itemize}
\end{itemize}

\checksubsection{Dataset Usage}
\begin{itemize}

\question{Does this paper rely on one or more datasets?}{(yes/no)}
yes

\ifyespoints{If yes, please address the following points:}
\begin{itemize}

	\question{A motivation is given for why the experiments are conducted on the selected datasets}{(yes/partial/no/NA)}
	yes

	\question{All novel datasets introduced in this paper are included in a data appendix}{(yes/partial/no/NA)}
	yes

	\question{All novel datasets introduced in this paper will be made publicly available upon publication of the paper with a license that allows free usage for research purposes}{(yes/partial/no/NA)}
	yes

	\question{All datasets drawn from the existing literature (potentially including authors' own previously published work) are accompanied by appropriate citations}{(yes/no/NA)}
	yes

	\question{All datasets drawn from the existing literature (potentially including authors' own previously published work) are publicly available}{(yes/partial/no/NA)}
	yes

	\question{All datasets that are not publicly available are described in detail, with explanation why publicly available alternatives are not scientifically satisficing}{(yes/partial/no/NA)}
	yes

\end{itemize}
\end{itemize}

\checksubsection{Computational Experiments}
\begin{itemize}

\question{Does this paper include computational experiments?}{(yes/no)}
yes

\ifyespoints{If yes, please address the following points:}
\begin{itemize}

	\question{This paper states the number and range of values tried per (hyper-) parameter during development of the paper, along with the criterion used for selecting the final parameter setting}{(yes/partial/no/NA)}
	yes

	\question{Any code required for pre-processing data is included in the appendix}{(yes/partial/no)}
	yes

	\question{All source code required for conducting and analyzing the experiments is included in a code appendix}{(yes/partial/no)}
	yes

	\question{All source code required for conducting and analyzing the experiments will be made publicly available upon publication of the paper with a license that allows free usage for research purposes}{(yes/partial/no)}
	yes
        
	\question{All source code implementing new methods have comments detailing the implementation, with references to the paper where each step comes from}{(yes/partial/no)}
	yes

	\question{If an algorithm depends on randomness, then the method used for setting seeds is described in a way sufficient to allow replication of results}{(yes/partial/no/NA)}
	yes

	\question{This paper specifies the computing infrastructure used for running experiments (hardware and software), including GPU/CPU models; amount of memory; operating system; names and versions of relevant software libraries and frameworks}{(yes/partial/no)}
	yes

	\question{This paper formally describes evaluation metrics used and explains the motivation for choosing these metrics}{(yes/partial/no)}
	yes

	\question{This paper states the number of algorithm runs used to compute each reported result}{(yes/no)}
	yes

	\question{Analysis of experiments goes beyond single-dimensional summaries of performance (e.g., average; median) to include measures of variation, confidence, or other distributional information}{(yes/no)}
	yes

	\question{The significance of any improvement or decrease in performance is judged using appropriate statistical tests (e.g., Wilcoxon signed-rank)}{(yes/partial/no)}
	yes

	\question{This paper lists all final (hyper-)parameters used for each model/algorithm in the paper’s experiments}{(yes/partial/no/NA)}
	yes

\end{itemize}
\end{itemize}

\end{document}